\pdfoutput=1
\documentclass[conference]{IEEEtran}
\usepackage{amsmath,amssymb,amsfonts}
\usepackage{pgfplots}
\pgfplotsset{compat=1.17}
\usetikzlibrary{shapes}
\usetikzlibrary{automata, arrows}
\usetikzlibrary{fit,positioning}
\usetikzlibrary{backgrounds}

\begin{document}

\title{Path Based Hierarchical Clustering on Knowledge Graphs}

\author{\IEEEauthorblockN{Marcin Pietrasik\IEEEauthorrefmark{1}, Marek Reformat\IEEEauthorrefmark{1}\IEEEauthorrefmark{2}}
\IEEEauthorblockA{{\IEEEauthorrefmark{1}Department of Electrical and Computer Engineering},
{University of Alberta},
Edmonton, Canada \\
}
\IEEEauthorblockA{\IEEEauthorrefmark{2}{Information Technology Institute},
{University of Social Sciences},
\L{}\'{o}d\'{z}, Poland\\
\{pietrasi, reformat\}@ualberta.ca}
}
\maketitle

\begin{abstract}
Knowledge graphs have emerged as a widely adopted medium for storing relational data, making methods for automatically reasoning with them highly desirable. In this paper, we present a novel approach for inducing a hierarchy of subject clusters, building upon our earlier work done in taxonomy induction. Our method starts with constructing a tag hierarchy before assigning subjects to clusters on this hierarchy. We quantitatively demonstrate our method's ability to induce a coherent cluster hierarchy on three real-world datasets.
\end{abstract}


\section{Introduction}
The widespread use of large-scale knowledge bases such as DBpedia \protect\cite{lehmann2015dbpedia} and Freebase \cite{bollacker2008freebase} has sparked a demand for ways of automatically reasoning with knowledge graphs in a scalable way. One aspect of this is discovering structure in the knowledge graph's data, such as inducing a hierarchy of subject clusters. This hierarchy would provide a summary of the information stored in the knowledge graph, aiding in its interpretability.
For instance, in a knowledge graph that describes films, a cluster hierarchy connecting the clusters \textit{comedy} and \textit{romantic comedy} would show that romantic comedies are of the type comedy as well as which films belong to each cluster.

In this paper, we present an extension to our method for inducing class taxonomies from knowledge graphs, \textit{SMICT} \cite{pietrasik2020simple}, which allows for inducing cluster hierarchies of knowledge graph subjects. The proposed approach is scalable to large-scale knowledge graphs and is able to induce coherent cluster hierarchies on three real-world datasets. In the remainder of this paper, we briefly discuss the related work before describing our method and evaluating it quantitatively.

\section{Related Work}

In an early method, \cite{roy2007learning} used a stochastic generative model to build a tree describing entity similarity.
\cite{nickel2012factorizing} performed hierarchical clustering on representations learned by \textit{RESCAL} \cite{nickel2011three} which factorizes a knowledge graph thereby generating features for its subjects, relations, and objects.
In an approach which bears similarity to our own, \cite{chen2014learning} describes each subject in a knowledge graph by its relation-object pairs. It then uses these pairs to calculate a similarity matrix between subjects on which agglomerative hierarchical clustering is performed using the extended Ward's minimum variance \cite{szekely2007measuring} as its measure.
\cite{mohamed2019unsupervised} takes a reversed approach wherein subjects which are described by the same relation-object pairs are assigned to the same groups. The similarity between these groups is then calculated to construct a hierarchy.

Taxonomy induction from knowledge graphs is a related task which involves constructing a taxonomy of a knowledge graph's type objects. Recently, we proposed the \textit{SMICT} method for this purpose which relies on object frequencies and co-occurrences, drawing inspiration from \cite{heymann2006collaborative} and \cite{schmitz2006inducing}. \textit{Statistical Schema Induction} \cite{volker2011statistical}, on the other hand, uses association rule mining on a knowledge graph's transaction table to induce taxonomy axioms. Entity and text embeddings are used for taxonomy induction in \textit{TIEmb} \cite{ristoski2017large} on the intuition that subclasses will be embedded within the radius of their superclasses in the embedding space.

\section{Approach}

Our approach is divided into three steps: hierarchy induction, subject clustering, and hierarchy pruning. In the first step, a tag hierarchy is induced from the knowledge graph using the aforementioned \textit{SMICT} method and defined by a set of subsumption axioms. This hierarchy serves as the base structure for the cluster hierarchy to which subjects are assigned in the second step. The final step prunes the hierarchy of empty clusters.

\subsection{Preliminaries}

We define a knowledge graph, $\mathcal{K}$, as a set of triples such that each triple relates a subject, $s \in \mathcal{S}$, to an object, $o \in \mathcal{O}$, via a relation, $r \in \mathcal{R}$. The knowledge graph may then be formalized as $\mathcal{K} = \{\langle s,r,o \rangle \in \mathcal{S} \times \mathcal{R} \times \mathcal{O}\}$.

\subsection{Hierarchy Induction}

\textit{SMICT} first transforms the knowledge graph's triple structure into subject-tag pairs where each tag, $t$, consists of a relation and object. Formally, $\mathcal{K} = \{ \langle s, t \rangle \in \mathcal{S} \times \mathcal{V} \}$ where tags are defined as $t := \langle r, o \rangle$ and $\mathcal{V}$ is the set of all tags. Furthermore, the knowledge graph is flattened on a single relation which describes the subject's type information.
The following statistics are calculated on the knowledge graph and serve as \textit{SMICT}'s input:
\begin{itemize}

\item The number of subjects annotated by tag $t_a$ is denoted as $\text{N}_{t_a}$.
\item The number of subjects annotated by both tags $t_a$ and $t_b$ is denoted as $\text{N}_{t_a, t_b}$.
\item The generality of tag $t_a$, denoted as $\text{G}_{t_a}$, is defined as:
\begin{equation}
    \text{G}_{t_a} = {\sum_{t_b \in \mathcal{V}_{-t_a}} \dfrac{\text{N}_{t_a,t_b}}{\text{N}_{t_b}}}
\end{equation}
where $\mathcal{V}_{-t_a}$ is the set of all tags excluding tag $t_a$.

\end{itemize}

First, the tag with the highest generality is initialized as the root of the hierarchy. Tags are then added greedily in order of decreasing generality as children of tags which have already been placed on the hierarchy. Parent tags are chosen as those that have the highest similarity with the tag being added. Similarity between tags $t_a$ and $t_b$, denoted as $\text{S}_{t_a \rightarrow t_b}$, is calculated as follows:
\begin{equation}
    \label{eqn:similarity}
    \text{S}_{t_a \rightarrow t_b} = \sum_{t_c \in \mathcal{P}_{t_a}} \alpha^{l_a - l_c}\dfrac{\text{N}_{t_b,t_c}}{\text{N}_{t_b}}
\end{equation}
where $\mathcal{P}_{t_a}$ is the set of tags in the path from the root tag to tag $t_a$. $l_a$ and $l_c$ denote the levels in the hierarchy of tags $t_a$ and $t_c$, respectively.  The decay factor, $\alpha$, is a hyperparameter that controls the effect ancestors of tag $t_a$ have on its similarity when calculating $\text{S}_{t_a \rightarrow t_b}$. For more information about this procedure, we refer readers to our original work \cite{pietrasik2020simple}.

\subsection{Subject Clustering}

The induced hierarchy is used to initialize the clusters such that each tag in the hierarchy becomes a cluster and the hierarchical relations between tags are extended to the clusters. The tags may then be seen as annotations for each cluster. We exploit this in our notation such that $c_a$ is the cluster initialized from tag $t_a$. Furthermore we denote $\mathcal{A}_i$ to be the set of all tags which annotate subject $s_i$.

Subjects are assigned to clusters by the degree to which they belong to a cluster. Belonging of subject $s_i$ to cluster $c_a$, denoted $\text{B}_{s_i \rightarrow c_a}$, is calculated as the Jaccard coefficient between the subject's tags, $\mathcal{A}_i$, and the tags encountered in the path from the root cluster to cluster $c_a$, denoted $\mathcal{P}_{c_a}$. Formally, this is calculated as:
\begin{equation}
    \text{B}_{s_i \rightarrow c_a} = \dfrac{|\mathcal{A}_i \cap \mathcal{P}_{c_a}|}  {|\mathcal{A}_i \cup \mathcal{P}_{c_a}|}
\end{equation}
Each subject is added to the cluster to which it has the highest degree of belonging. This process may be parallelized to increase performance.

\subsection{Hierarchy Pruning}

The two previous steps may induce a hierarchy containing empty clusters which need to get pruned. Pruning is performed by traversing the hierarchy depth first and removing all empty clusters. In addition, non-empty clusters which have empty parent clusters are reattached as the children of their first non-empty ancestor. If a non-empty cluster has no non-empty ancestors, it becomes the child of the root. The root cluster is never removed, regardless of whether it is empty or not.

\section{Evaluation}

A Python implementation of our method as well as the datasets used in our evaluation are publicly available on Github\footnote{https://github.com/mpietrasik/smich}.

\subsection{Evaluation Procedure}

We evaluate our method on three real-world datasets: IIMB, DBpedia, and WordNet.
\begin{itemize}
 
\item The IIMB dataset \cite{euzenat2011results} was created from Freebase for the 2010 Ontology Alignment Evaluation Initiative. We added a root tag to all subjects to anchor the hierarchy. The dataset contains 1416 subjects and 82 tags.

\item The DBpedia dataset \cite{pietrasik2020simple} was generated from a subset of DBpedia and contains 50,000 subjects and 418 tags.

\item The WordNet dataset \cite{pietrasik2020simple} was generated by querying DBpedia for subjects which appear in WordNet \cite{miller1995wordnet} and contains 50,000 subjects and 1752 tags.
\end{itemize}

Performance is evaluated by calculating the $F_1$ score \cite{van1979information} of: the induced tag hierarchy (Hie-$F_1$); the belonging of subjects to their clusters (Sub-$F_1$); and how well clusters represent the tags in the vocabulary (Tag-$F_1$).
The Hie-$F_1$ score is obtained by comparing the induced subsumption axioms against a gold standard hierarchy as described in \cite{pietrasik2020simple}. 
Sub-$F_1$ and Tag-$F_1$ highlight the trade-off between large, heterogeneous clusters on a strongly heritable hierarchy (favoured by Sub-$F_1$) and smaller homogeneous clusters on a less heritable hierarchy (favoured by Tag-$F_1$). 
For obtaining the former, each cluster inherits all the subjects of its descendant clusters and the $F_1$ score is calculated such that a subject is correctly assigned to a cluster if both subject and cluster are annotated by the same tag. 
The latter is obtained in a way similar to the technique used in \cite{nickel2012factorizing}. As before, each cluster inherits all the subjects of its descendant clusters and the $F_1$ score between each tag and each cluster is calculated. The $F_1$ that is highest among the clusters becomes the score of the tag.
To highlight hyperparameter sensitivity, we performed a sweep of $\alpha$ values $0 < \alpha < 1$ in increments of $0.05$.

\subsection{Results}

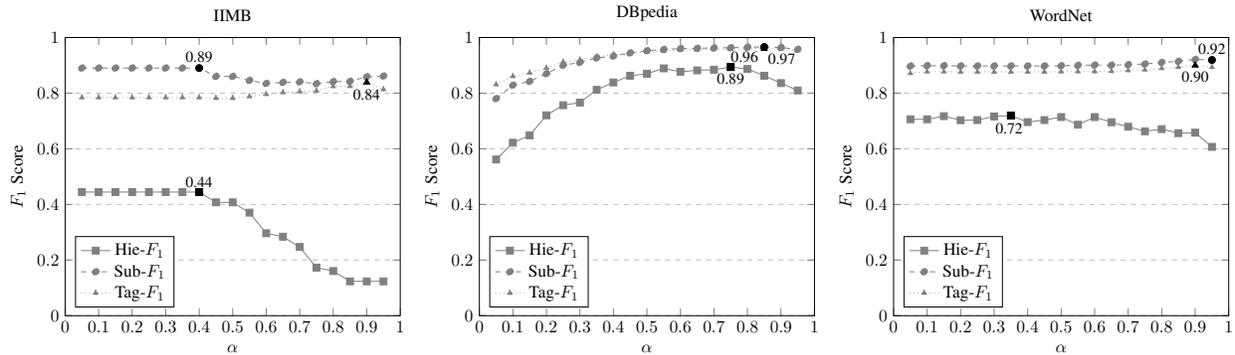
\begin{figure*}
\centering
\begin{tikzpicture}[scale=0.65]
\begin{axis}[
    title = { IIMB },
    xlabel={$\alpha$},
    ylabel={$F_1$ Score},
    xmin=0, xmax=1,
    ymin=0, ymax=1,
    log ticks with fixed point,
    xtick={0,0.1,0.2,0.3,0.4,0.5,0.6,0.70,0.8,0.9,1},
    ytick={0,0.2,0.4,0.6,0.8,1},
    legend pos=south west,
    ymajorgrids=true,
    grid style=dashed,
]
\addplot[color=gray, mark=square*,]
    coordinates {(0.05, 0.4444444444444444)(0.1, 0.4444444444444444)(0.15, 0.4444444444444444)(0.2, 0.4444444444444444)(0.25, 0.4444444444444444)(0.3, 0.4444444444444444)(0.35, 0.4444444444444444)(0.4, 0.4444444444444444)(0.45, 0.4074074074074074)(0.5, 0.4074074074074074)(0.55,
0.37037037037037035)(0.6, 0.2962962962962963)(0.65, 0.2839506172839506)(0.7, 0.24691358024691357)(0.75, 0.1728395061728395)(0.8, 0.16049382716049382)(0.85, 0.12345679012345678)(0.9, 0.12345679012345678)(0.95, 0.12345679012345678)};

\addplot[color=gray, dashed, mark=*,]
    coordinates {(0.05, 0.8902507168821207)(0.1, 0.8902507168821207)(0.15, 0.8902507168821207)(0.2, 0.8902507168821207)(0.25, 0.8902507168821207)(0.3, 0.8902507168821207)(0.35, 0.8902507168821207)(0.4, 0.8902507168821207)(0.45, 0.8602428034577794)(0.5, 0.8602428034577794)(0.55,
0.8458344977573267)(0.6, 0.8349023268493209)(0.65, 0.8386791457432883)(0.7, 0.8405316446887453)(0.75, 0.834273236463404)(0.8, 0.8412589364373771)(0.85, 0.8425164282661332)(0.9, 0.8589552018149702)(0.95, 0.8613937174634911)};
    
\addplot[color=gray, dotted, mark=triangle*,]
    coordinates {(0.05, 0.7842913158061362)(0.1, 0.7842913158061362)(0.15, 0.7842913158061362)(0.2, 0.7842913158061362)(0.25, 0.7842913158061362)(0.3, 0.7842913158061362)(0.35, 0.7842913158061362)(0.4, 0.7842913158061362)(0.45, 0.7820785416476593)(0.5, 0.7820785416476593)(0.55,
0.7880808183163162)(0.6, 0.7959256157652641)(0.65, 0.8024758356314813)(0.7, 0.8055405902844464)(0.75, 0.8086922872867022)(0.8, 0.8244468240734574)(0.85, 0.8250127051834134)(0.9, 0.8389644756453871)(0.95, 0.8136173732940324)};

\addplot[color=black, mark=square*,]
coordinates{(0.4, 0.4444444444444444)};
\node[black,above] at (0.40,0.44){\small{0.44}};

\addplot[color=black, mark=*,]
coordinates{(0.4, 0.8902507168821207)};
\node[black,above] at (0.40,0.8902){\small{0.89}};

\addplot[color=black, mark=triangle*,]
coordinates{(0.9, 0.8389644756453871)};
\node[black,below] at (0.90,0.8370){\small{0.84}};

\legend{Hie-$F_1$, Sub-$F_1$, Tag-$F_1$}
\end{axis}
\end{tikzpicture}
\begin{tikzpicture}[scale=0.65]
\begin{axis}[
    title = { DBpedia },
    xlabel={$\alpha$},
    ylabel={$F_1$ Score},
    xmin=0, xmax=1,
    ymin=0, ymax=1,
    log ticks with fixed point,
    xtick={0,0.1,0.2,0.3,0.4,0.5,0.6,0.70,0.8,0.9,1},
    ytick={0,0.2,0.4,0.6,0.8,1},
    legend pos=south west,
    ymajorgrids=true,
    grid style=dashed,
]
\addplot[color=gray, mark=square*,]
    coordinates {(0.05, 0.5614457831325301)(0.1, 0.6216867469879518)(0.15, 0.6481927710843375)(0.2, 0.7204819277108434)(0.25, 0.7566265060240963)(0.3, 0.7662650602409637)(0.35, 0.8120481927710843)(0.4, 0.83855421686747)(0.45, 0.8626506024096385)(0.5, 0.8698795180722891)(0.55, 0.8891566265060241)(0.6, 0.8771084337349399)(0.65, 0.8819277108433735)(0.7, 0.8843373493975902)(0.75, 0.8939759036144579)(0.8, 0.8867469879518073)(0.85, 0.8626506024096385)(0.9, 0.8361445783132531)(0.95, 0.8096385542168675)};

\addplot[color=gray, dashed, mark=*,]
    coordinates {(0.05, 0.7803128762972704)(0.1, 0.8286815595694184)(0.15, 0.842474001801363)(0.2, 0.8699436110421727)(0.25, 0.90008165712285)(0.3, 0.9106228301429198)(0.35, 0.9264552351659998)(0.4, 0.9331007911016413)(0.45, 0.9446462525383643)(0.5, 0.9527662325913681)(0.55, 0.9570066477570548)(0.6, 0.9602170267646052)(0.65, 0.9612163073217507)(0.7, 0.9624317000455421)(0.75, 0.9631888592763669)(0.8, 0.9653435816691195)(0.85, 0.9659285553128993)(0.9, 0.9639473573260182)(0.95, 0.9576075618742359)};
    
\addplot[color=gray, dotted, mark=triangle*,]
    coordinates {(0.05, 0.831052041247756)(0.1, 0.8607337221086974)(0.15, 0.8731013333964469)(0.2, 0.8902604168828465)(0.25, 0.9124945499449896)(0.3, 0.9197792684551352)(0.35, 0.9310200974549181)(0.4, 0.9381183080387081)(0.45, 0.9425714027288356)(0.5, 0.9480524379113873)(0.55, 0.9514561741015504)(0.6, 0.9544822251278061)(0.65, 0.9562323596336922)(0.7, 0.9571601753989261)(0.75, 0.9578756727581983)(0.8, 0.9599118338709669)(0.85, 0.9602807614386308)(0.9, 0.9589363195746676)(0.95, 0.9529426512209064)};
    
\addplot[color=black, mark=square*,]
coordinates{(0.75,0.8939759036144579)};
\node[black,below] at (0.75,0.8939){\small{0.89}};

\addplot[color=black, mark=*,]
coordinates{(0.85, 0.9659285553128993)};
\node[black,below right] at (0.844, 0.9659){\small{0.97}};

\addplot[color=black, mark=triangle*,]
coordinates{(0.85, 0.9602807614386308)};
\node[black,below left] at (0.85,0.9722){\small{0.96}};

\legend{Hie-$F_1$, Sub-$F_1$, Tag-$F_1$}
\end{axis}
\end{tikzpicture}
\begin{tikzpicture}[scale=0.65]
\begin{axis}[
    title = { WordNet },
    xlabel={$\alpha$},
    ylabel={$F_1$ Score},
    xmin=0, xmax=1,
    ymin=0, ymax=1,
    log ticks with fixed point,
    xtick={0,0.1,0.2,0.3,0.4,0.5,0.6,0.70,0.8,0.9,1},
    ytick={0,0.2,0.4,0.6,0.8,1},
    legend pos=south west,
    ymajorgrids=true,
    grid style=dashed,
]
\addplot[color=gray, mark=square*,]
    coordinates {(0.05, 0.7061132922041503)(0.1, 0.7061132922041503)(0.15, 0.7173303421200224)(0.2, 0.7027481772293886)(0.25, 0.7033090297251823)(0.3, 0.7162086371284352)(0.35, 0.7195737521031967)(0.4, 0.6960179472798654)(0.45, 0.7033090297251823)(0.5, 0.7139652271452607)(0.55, 0.6864834548513741)(0.6, 0.7139652271452607)(0.65, 0.6954570947840717)(0.7, 0.6791923724060571)(0.75, 0.6629276500280425)(0.8, 0.6707795849691531)(0.85, 0.6561974200785194)(0.9, 0.6578799775659001)(0.95, 0.606842400448682)};

\addplot[color=gray, dashed, mark=*,]
    coordinates {(0.05, 0.8973446735459668)(0.1, 0.8986911404911772)(0.15, 0.8986911404911772)(0.2, 0.8977439772966631)(0.25, 0.8977439772966631)(0.3, 0.8977439772966631)(0.35, 0.8977439772966631)(0.4, 0.8977439772966631)(0.45, 0.8977439772966631)(0.5, 0.8985828219954693)(0.55, 0.9000912694980061)(0.6, 0.9010328100532339)(0.65, 0.9006614304978902)(0.7, 0.9024445434714615)(0.75, 0.9048042276244928)(0.8, 0.9105776825574027)(0.85, 0.9145942239213682)(0.9, 0.9212113088918458)(0.95, 0.9192247899354468)};
    
\addplot[color=gray, dotted, mark=triangle*,]
    coordinates {(0.05, 0.871616837080615)(0.1, 0.8772612531810521)(0.15, 0.8772612531810521)(0.2, 0.8764548474185706)(0.25, 0.8764548474185706)(0.3, 0.8764548474185706)(0.35, 0.8764548474185706)(0.4, 0.8764548474185706)(0.45, 0.8764548474185706)(0.5, 0.8771919894169173)(0.55, 0.8774886897115715)(0.6, 0.8778462328223154)(0.65, 0.8787610151595382)(0.7, 0.881759760843255)(0.75, 0.8838236290401109)(0.8, 0.8892964978931005)(0.85, 0.8939347369693172)(0.9, 0.8998483312018625)(0.95, 0.8932236034995225)};
    
\addplot[color=black, mark=square*,]
coordinates{(0.35,0.7195737521031967)};
\node[black,below] at (0.345,0.715){\small{0.72}};

\addplot[color=black, mark=*,]
coordinates{(0.95, 0.9192247899354468)};
\node[black,above] at (0.95, 0.919){\small{0.92}};

\addplot[color=black, mark=triangle*,]
coordinates{(0.9, 0.8998483312018625)};
\node[black,below] at (0.90,0.8998){\small{0.90}};

\legend{Hie-$F_1$, Sub-$F_1$, Tag-$F_1$}
\end{axis}
\end{tikzpicture}

\caption{Results obtained on the three datasets. Black highlight indicates highest performance for each metric.}
\label{fig:results}
\end{figure*}

The results of the aforementioned metrics on the three datasets are summarized in Figure \ref{fig:results}. We do not notice a consistent pattern between metrics. This is expected since they each measure different aspects of the hierarchy and optimal performance across all metrics may not be possible at the same $\alpha$ value. 
At Sub-$F_1$ scores near or above 0.9, we conclude that subjects are assigned to correct clusters at a very high rate. Moreover, high Tag-$F_1$ scores indicate a high degree of cluster separation and homogeneity.
A high discrepancy between Hie-$F_1$ and the other two metrics indicates a dissimilarity between the structure of the gold standard hierarchy and how subjects are annotated in the knowledge graph.

\begin{figure*}
\centering
\begin{tikzpicture}[scale = 0.96, sibling distance=9.5em,
  every node/.style = {shape=rectangle, rounded corners,
    draw, align=justify,
    top color=white, bottom color=white}]]
  \node {\textbf{\footnotesize{root}}}
      child {  node [rectangle split, rectangle split, rectangle split parts=2,
         text ragged, font=\scriptsize\selectfont] {
            \textbf{\footnotesize{location}}
                  \nodepart{second}
                   \scriptsize{anhui burnaby} \\ \scriptsize{miami queens}
        }
    child {  node [rectangle split, rectangle split, rectangle split parts=2,
         text ragged, font=\scriptsize\selectfont, anchor=north] {
            \textbf{\footnotesize{country}}
                  \nodepart{second}
                   \scriptsize{canada colombia} \\ \scriptsize{germany scotland}
        }
    }
    child {  node [rectangle split, rectangle split, rectangle split parts=2,
         text ragged, font=\scriptsize\selectfont, anchor=north] {
            \textbf{\footnotesize{city}}
                  \nodepart{second}
                   \scriptsize{havana madrid} \\ \scriptsize{montevideo prague}
        }
    }
    }
    child {  node [rectangle split, rectangle split, rectangle split parts=2,
         text ragged, font=\scriptsize\selectfont, anchor=north] {
            \textbf{\footnotesize{language}}
                  \nodepart{second}
                   \scriptsize{english french} \\ \scriptsize{polish russian}
        }
    }
    child {  node [rectangle split, rectangle split, rectangle split parts=2,
         text ragged, font=\scriptsize\selectfont, anchor = north] {
            \textbf{\footnotesize{film}}
                  \nodepart{second}
                   \scriptsize{seven\_swords shane} \\ \scriptsize{some\_girls\_do spy\_game}
        }
        child { node [rectangle split, rectangle split, rectangle split parts=2,
         text ragged, text ragged, font=\scriptsize\selectfont] {\textbf{\footnotesize{comedy}}
         \nodepart{second}
         \scriptsize{schtonk scoop } \\ \scriptsize{silverado strange\_brew} }
            child { node [rectangle split, rectangle split, rectangle split parts=2,
         text ragged, text ragged, font=\scriptsize\selectfont,anchor = east] {
            \textbf{\footnotesize{musical}}
                  \nodepart{second}
                   \scriptsize{school\_of\_rock seven\_brides\_for\_seven\_brothers} \\ \scriptsize{singin\_in\_the\_rain south\_park\_bigger\_longer\_uncut}
        } }
        child { node [rectangle split, rectangle split, rectangle split parts=2,
         text ragged, text ragged, font=\scriptsize\selectfont, anchor = west] {
            \textbf{\footnotesize{buddy film}}
                  \nodepart{second}
                   \scriptsize{shanghai\_knights stripes}  \\ \scriptsize{starsky\_hutch\_2004 swingers}
        }}
        }
       }
     child {  node [rectangle split, rectangle split, rectangle split parts=2,
         text ragged, text ragged, font=\scriptsize\selectfont,  anchor = north] {
            \textbf{\footnotesize{actor}}
                  \nodepart{second}
                   \scriptsize{james\_woods brad\_pitt} \\ \scriptsize{meg\_ryan tom\_hanks} 
        }
    }
    child {  node [rectangle split, rectangle split, rectangle split parts=2,
         text ragged, text ragged, font=\scriptsize\selectfont] {
            \textbf{\small{director}}
                  \nodepart{second}
                   \scriptsize{alfred\_hitchcock andrei\_tarkovsky}\\ \scriptsize{richard\_linklater stanley\_kubrick}
        }
        child { node [rectangle split, rectangle split, rectangle split parts=2,
         text ragged, text ragged, font=\scriptsize\selectfont, anchor=north] {
            \textbf{\footnotesize{character creator}}
                  \nodepart{second}
                   \scriptsize{george\_lucas stanislaw\_lem} \\ \scriptsize{stan\_lee steve\_ditko}
        } }
    }
      ;
\end{tikzpicture}
\caption{Excerpt of the cluster hierarchy induced on the IIMB dataset. Node top indicates cluster's tag; bottom indicates cluster's constituent subjects.}
\label{fig:hierarchy}
\end{figure*}
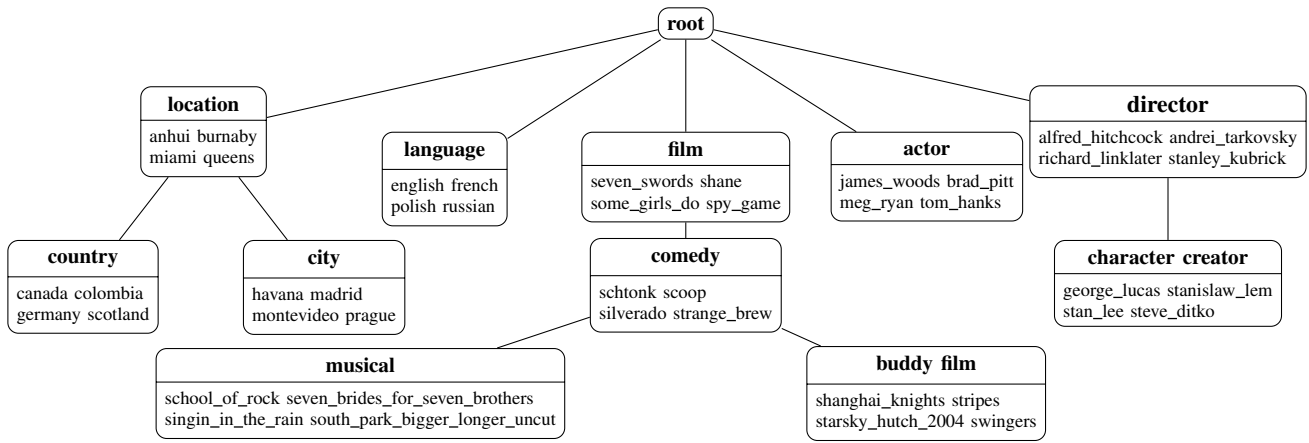

Figure \ref{fig:hierarchy} provides an excerpt of the cluster hierarchy induced on the IIMB dataset at $\alpha = 0.9$. The full cluster hierarchies for each dataset are available on Github.

\section{Conclusion}

In this paper we proposed an extension to \textit{SMICT} that allows for inducing cluster hierarchies of knowledge graph subjects. Our approach was evaluated on three real-world datasets and shown to construct coherent cluster hierarchies as per our evaluation metrics. The code, datasets, and results used in our work have been published online for replication.

\bibliographystyle{IEEEtran}
\bibliography{bibliography}

\end{document}